# Spectral Analysis of Projection Histogram for Enhancing Close matching character Recognition in Malayalam

*Sajilal Divakaran*
University of Kerala, Thiruvananthapuram,
Kerala, India 695581
sajilald@gmail.com

*Abstract*

*The success rates of Optical Character Recognition (OCR) systems for printed Malayalam documents is quite impressive with the state of the art accuracy levels in the range of 85-95% for various. However for real applications, further enhancement of this accuracy levels are required. One of the bottle necks in further enhancement of the accuracy is identified as close-matching characters. In this paper, we delineate the close matching characters in Malayalam and report the development of a specialised classifier for these close-matching characters. The output of a state of the art of OCR is taken and characters falling into the close-matching character set is further fed into this specialised classifier for enhancing the accuracy. The classifier is based on support vector machine algorithm and uses feature vectors derived out of spectral coefficients of projection histogram signals of close-matching characters.*

*Keywords* - OCR, Malayalam, Close-matching Characters, Feature Extraction, Pattern Classification

## I. INTRODUCTION

With the advent of internationalisation in computing, every world language is being integrated into digital technology in a very fine way. The progress is naturally dependent on the language user base. For languages which do not have in critical mass of user base, the efforts are naturally limited due to the minimality of the business scope. Malayalam, the official language of the state of Kerala, is an example of such a case. Malayalam belongs to the Dravidian family of languages and is one of the four major languages of this family. The native language of the South Indian state of Kerala and Lakshadweep Islands in the west coast of India, Malayalam is spoken by about 4% of India's population.

On a very basic level, Malayalam language has been integrated well into the digital technology. Unicode fonts for Malayalam are available for desktop publishing software. Both Unicode and non-Unicode fonts are widely available, though with varying keyboard support which is yet to be standardised in a practical sense. More involved technologisation such as speech recognition, synthesis, character recognition is underway with varying degrees of success. Research and development efforts by the Centre for Development of Advanced Computing(C-DAC) have resulted in relatively successful text-to-speech system in Malayalam named *Subhashini* [1]. Layout Analyzer and Optical Character Recognition system for printed Malayalam Documents, *Nayana* is also a product of C-DAC [2]. However there is scope for both fine and coarse improvement in these systems and also enormous scope for taking up unattended technologisation issues in Malayalam. In this paper we address a fine improvement in the existing OCR systems by developing a specialized pattern classifier for close-matching characters of Malayalam, and thereby providing a means of enhancing the character recognition of accuracy of OCR systems for Malayalam.

| Close matching Character | | Recognition Error (%) |
|---|---|---|
| ) | ാ | 100 |
| ( | ၂ | 100 |
| ച | ഖ | 16.66 |
| ക്ക | ക്ങ | 29.33 |
| ക | കു | 45.238 |
| പൂ | പ്ല | 66.66 |
| ? | ാ | 14.28 |
| ? | ി | 14.28 |
| ു | ൂ | 50 |
| ത | ങ | 24.59 |
| ത്ത | ങ്ങ | 100 |

Table. 1 Close-matching character in Malayalam that reduces recognition rates of state of the art *Nayana* [2].

The Malayalam character set consists of 51 letters which includes 13 vowels and 37 consonants. The set also consists of 12 vowel signs. Table. 1 shows the close-matching characters in the Malayalam character set that produce accuracies less than 10% in *Nayana,* state-of-the art OCR for Malayalam. A specialised classifier for the above characters is proposed to be integrated into the standard OCR system as shown in Fig 1.

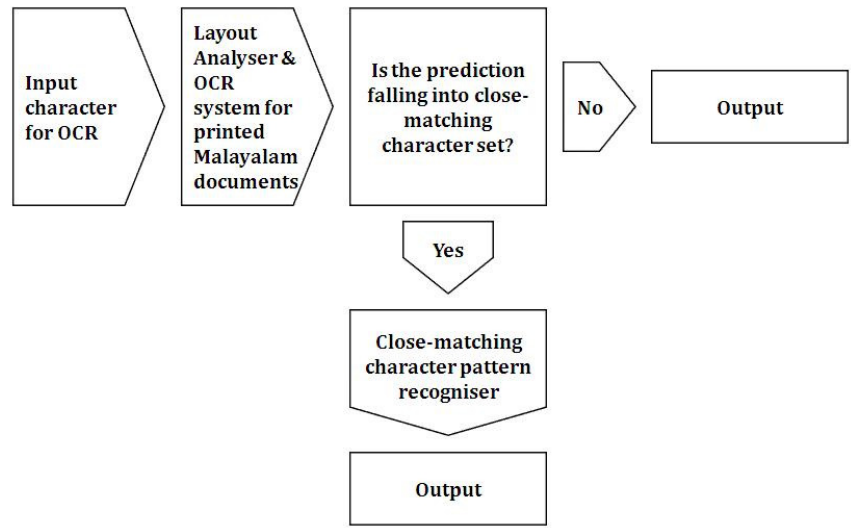

Fig. 1 Layout of the proposed closed-matching System

## II. Malayalam OCR Systems

We briefly describe the state of the art OCR system for Malayalam which is similar to other world languages too. It consists of the following phases: Preprocessing, Recognition and Post processing. In the preprocessing section, the text document is scanned and a bit mapped image is produced. This grey-level image is converted into a black and white binary images and standard noise removal algorithms are applied and skew corrections is also done on the image if necessary. Page segmentation and block classification are also done to finally arrive at character segmentation [3]. This part of the phase of preprocessing is referred to as layout analysis. Once the character segmentation is done, then the phase of pattern recognition for characters becomes active. The character recognition phase uses a tool trained on features extracted from a training set of characters which classifies every input character into a identified character code. For every new incoming character, the feature is extracted and then fed into this classifier which produces a draft text. The core of this recognizer is the selection of meaningful features that capture the essence of uniqueness of each character. The fine tuning of the pattern recognition tool, which are very often neural networks or support vector machine or other similar soft computing methods, is also crucial [4]. After the recognition by the classifier, post processing can be done in a variety of ways to enhance the accuracy level. The errors in the character can be corrected by using the linguistic rule of the language to identify some common impossible combination of character that might appear in the predictor. This goes a long way in enhancing the accuracy of the total system. However this is not assured to work in every case as there could be errors in character recognition that may not violate any linguistic rule. An example in English could be that of "of" and "if".

State of art Malayalam OCR system such as Nayana uses intelligent character recognition, or topological feature analysis. They base the recognition on features specific to each character. General features such as closed loop, vertical lines and points, etc. are used to identify individual characters. The character recognition module used by Nayana extracts these features from segmented characters and these features are input into a binary tree classifier [5]. After passing through the different stages of the classifier the character is identified and corresponding character code is assigned.

Nayana has selected features based on topology/end points and run numbers for the classification for of Malayalam characters. The topological and stroke based features are used for the initial classification of characters [6]. Their considerations in selecting the features have been character structure, reliability of the feature, font and size independence and speed of feature extraction. They have grouped their features into three types. Type 1 features consist of vertical left bar, vertical right bar, vertical mid bar, left part bar and right part bar and also horizontal bar and horizontal bottom bars. Type 2 features consist of loops, number of loops, loop height and loop positions, aspect ratio, horizontal and vertical cutting points. Type 3 features are end points, number and location of end points and run number based features. Type 1 and Type 2 are claimed to be independent of font size and style and thus form the main features for Nayana [2].

The binary tree classifier used in Nayana is a hierarchical classifier which compares incoming data with a range of features. The selection of features is done by an assessment of separability of the classes. The decision tree system is a deterministic system for the classification which has the advantage that the computing time is very less. However the accuracy of the system is dependent on the design of the decision tree. We perceive that this is the limiting factor for the maximum level of accuracy obtained by Nayana and hence the work reported in this paper has discarded with a binary tree classifier and chosen a soft computing approach instead.

## III. FEATURE VECTOR FOR CLOSE-MATCHING CHARACTERS

We consider the selected close matching character set (Table 1) in Malayalam and describe a classifier for that based on unique feature extraction. Feature extraction is a crucial stage in design of any pattern classifier. The objects to be classified or recognised (in this case character images) are often bulky and cannot be directly taken as input data to the classifier. Hence a limited set of powerful features are to be extracted which will faithfully represent the uniqueness of the object to be classified. The intention is to extract set of features that will maximise the recognition the rate. The selection of the features also needs to take it to consideration that there could be a high degree of variability due to either nature of handwriting or changes in fonts. Feature extraction is typically based on three types of features, statistical, structural and global transformations and moments. Statistical feature is the representation of a character image by statistical distribution of points and takes care of style variations to some extent [7]. The major statistical features used for character representation are zoning, projections & profiles and crossings & distances. Structural features have high tolerance to distortions and style variations. This type of representation may also encode some knowledge about the structure of the object or may provide some knowledge as to what sort of components make up that object. Structural features are based on topological and geometrical properties of the character, such as aspect ratio, cross points, loops, branch points, strokes and their directions, inflection between two points, horizontal curves at top or bottom, etc... Global Transformations and Moments make the process of recognition as scale, translation, and rotation invariant. The original image can be completely reconstructed from the moment coefficients.

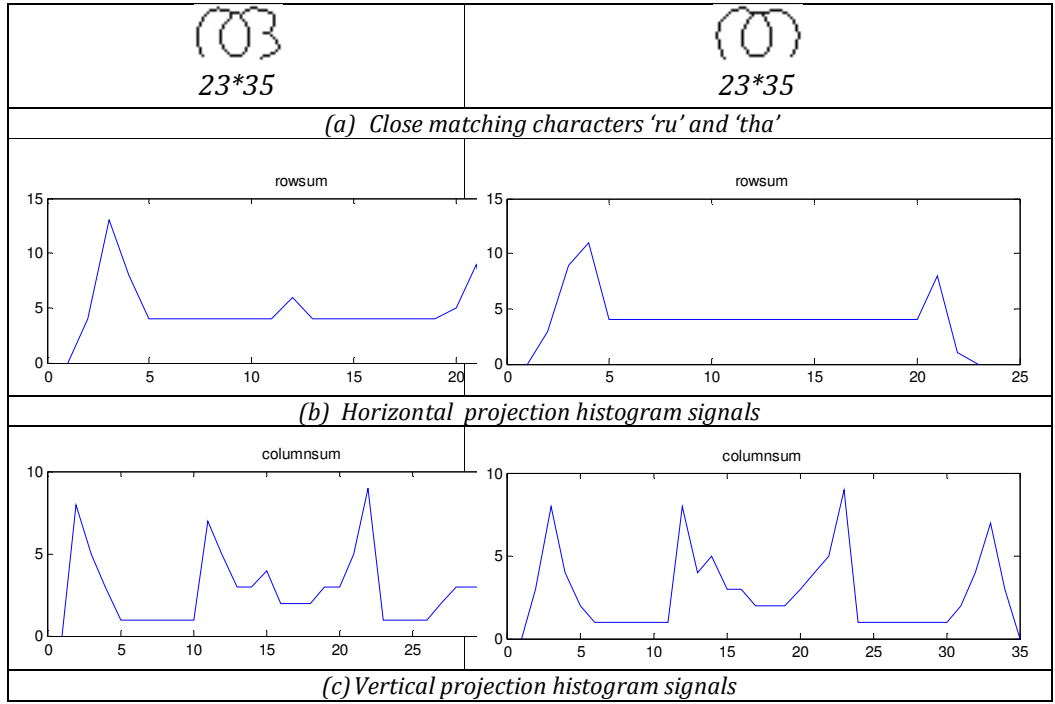

Fig. 2 Extracting Projection Histogram Signals for close matching characters – the case of '*ru*' and '*tha*'. Though the characters are seen as closely matching, the signals derived are very unique.

The use of the above features has however not been shown to be powerful enough in the case of close matching characters as evidenced by the accuracy levels reported for *Nayana* [2]. It has been pointed out the developers of *Nayana* that one of the causes of drop in accuracy levels is the confusion caused by close-matching characters in Malayalam. This observation is the motivation for this study.

We considered the possibility of a more objective extraction of features rather than going into the hand picking of features of each character and building a decision tree. We have therefore considered the character images disregarding the type of characters or the content of the character and we attempted to extract features which are unique. We report projection histograms as the means of extracting a unique and powerful feature vectors. A projection histogram is obtained by plotting the pixel distribution in horizontal and vertical directions of a character image. Fig 2. shows the projection histogram signals of two close matching characters in Malayalam. It obviously highlights the power of these signals to generate unique representations even when characters match closely.

The basic transformation that a projection histogram does is to convert the original images which are 2-D signals to one-dimensional signals. These signals are independent of noise and deformation but depend on the rotation [8]. Also, one dimensional signal can easily be processed using digital signal processing tools such as Discrete Fourier Transforms [9]. Fourier analysis is a well-known mathematical method by which any complex signal can be analysed as a sum of a large number of constituent sinusoidal signals [10]. The application of Discrete Fourier transforms would help to reduce the feature vector to any size of our choice, of course with the implications on accuracy. An n-point data set can be converted into an n-point DFT coefficient vector. It is well known that the higher coefficients in such vectors can be discarded and the signal built back with very good similarity. This indeed is our approach in extracting a feature vector. Consider an N-point projection histogram signal represented as a vector S(n). Its spectral content can be calculated using the following formula:

$$S(k) = \sum_{n=0}^{N-1} S(n) e^{-j2\pi kn/N}, k = 0, 1, 2, ...., N-1$$

S(k) is an n-point vector and can be reduced to any size of our choice by discarding higher coefficients. This is justifiable as high frequency changes in the character images are not a determinant for visual identification of characters. Suppose the horizontal and vertical histogram projection signals of an image are given by Sh and Sv, then the feature Vector X is extracted as:

$$X = [f_{h1}, f_{h2}, ...., f_{hM}; f_{v1}, f_{v2}, ,.... f_{vM}]$$

Where fhi is the i-th Fourier coefficient for the horizontal projection histogram and fvi is the i-th Fourier coefficient of the vertical projection histogram. The image size is considered as N x N, and M ≤ N.

## IV. RESULTS, DISCUSSIONS AND CONCLUSION

We utilised the training set used in Nayana system for the training of the selected close-matching character recogniser. We first separated out the close matching characters training set from the full data set and divided it into two equal training and test data sets with an extracted the projected histogram in the X & Y directions. We computed the Discrete Fourier Transform (DFT) coefficients and discarded half of the coefficient and formed a feature vector for the each of the close-matching character set. We used a Support Vector Machine (SVM) [11] for developing the classifier tool. SVMs use an enhancement of the n-dimensional plane separation that is achieved by traditional artificial neural networks. Instead of the simple linear separation, the maximal band of separation between the pattern clusters is achieved and hence SVMs are generally superior to the other soft computing methods.

We used the LIBSVM software [11] to develop our system. We choose the RBF Kernel [12] due to its simplicity and its ability to handle nonlinear class-separation boundaries. We show below in Table 2. the recognition accuracy of our trained SVM for characters which gave less than 10% accuracy in Nayana. Our results are indeed a major improvement for the state of the art. Table 2 clearly implies that merging the result of the specialised close-matching character recogniser with the original Nayana system will enhance the overall prediction accuracy of the latter.

| Correct Character | Error Character | Sensitivity | Specificity | Accuracy |
|---|---|---|---|---|
| ) | ാാ | 100 | 58.33 | 79.167 |
| ( | ㄴ | 75 | 100 | 87.5 |
| ഛ | ഖ | 100 | 41.67 | 70.83 |
| ക്ന | ക്ദ | 100 | 41.67 | 70.83 |
| ക | ക്ദ | 100 | 41.67 | 70.83 |
| പ്പ | പ്ല | 100 | 41.67 | 70.83 |
| ? | ാാ | 100 | 41.67 | 70.83 |
| ? | ി | 41.67 | 100 | 70.83 |
| ു | ൂ | 100 | 66.67 | 83.33 |
| ന | ഌ | 100 | 41.67 | 70.83 |
| ൊ | ഌ | 91.67 | 50 | 70.83 |

Table 2. Recognition Accuracy in the close-matching character recogniser for Malayalam.

Switching from a subjective feature extraction into an objective signal processing based feature extraction could perhaps be extended to the whole character set. To achieve further accuracy, we feel that the transformation of the image into more unique representation has potential to be a very powerful objective feature extraction. A candidate that currently we are investigating is the RQA (Recurrence Quantification Analysis) [13] for such transformations. We also feel that there is a need to re-open the scope of the OCR in Malayalam to move from text documents to identification of characters from still and video images, a field to the best of our knowledge, is yet to be addressed.

## ACKNOWLEDGEMENT

The authors gratefully acknowledges *Mrs K G Sulochana,* Joint Director, Centre for Development of Advanced Computing (C-DAC) for providing required information about *Nayana* and offering valuable inputs through discussions.